\documentclass[journal]{IEEEtran}

\usepackage{amsmath,amsfonts}
\usepackage{algorithm}
\usepackage{algorithmic}
\usepackage{array}
\usepackage{textcomp}
\usepackage{stfloats}
\usepackage{url}
\usepackage{verbatim}
\usepackage{graphicx}
\usepackage{tikz}
\usetikzlibrary{arrows.meta,positioning,fit,calc}
\usetikzlibrary{quantikz2}
\tikzset{
	qgate h/.style={draw={rgb,255:red,191;green,60;blue,60}, fill={rgb,255:red,255;green,235;blue,235}, text={rgb,255:red,145;green,35;blue,35}, rounded corners=2pt, line width=0.6pt},
	qgate rx/.style={draw={rgb,255:red,42;green,92;blue,180}, fill={rgb,255:red,232;green,240;blue,255}, text={rgb,255:red,20;green,62;blue,130}, rounded corners=2pt, line width=0.55pt},
	qgate ry/.style={draw={rgb,255:red,45;green,145;blue,96}, fill={rgb,255:red,232;green,248;blue,239}, text={rgb,255:red,24;green,105;blue,70}, rounded corners=2pt, line width=0.55pt},
	qgate rz/.style={draw={rgb,255:red,126;green,82;blue,178}, fill={rgb,255:red,242;green,235;blue,250}, text={rgb,255:red,92;green,54;blue,142}, rounded corners=2pt, line width=0.55pt},
	qgate cnot/.style={draw={rgb,255:red,208;green,105;blue,41}, fill={rgb,255:red,255;green,242;blue,226}, text={rgb,255:red,156;green,71;blue,26}, line width=0.6pt},
	qwire cnot/.style={draw={rgb,255:red,208;green,105;blue,41}, line width=0.6pt},
	qgate measure/.style={draw={rgb,255:red,112;green,118;blue,126}, fill={rgb,255:red,239;green,241;blue,244}, text={rgb,255:red,74;green,79;blue,87}, rounded corners=2pt, line width=0.55pt}
}
\usepackage[export]{adjustbox}% for the second solution
\usepackage{ntheorem}
\usepackage{longtable}
\usepackage{booktabs}
\usepackage{makecell}
\usepackage{amsmath}
\usepackage{graphicx}
\usepackage{mwe}
\usepackage{graphbox}
\usepackage{amsfonts} 
\usepackage{enumitem}
\usepackage{amssymb}
\usepackage{ragged2e}
\usepackage[colorlinks, linkcolor=blue, anchorcolor=blue, citecolor=blue]{hyperref}%为参考文献设置超链接
\usepackage[numbers,sort&compress]{natbib}%针对参考文献合并显示
\usepackage{comment}
\usepackage{color}
\usepackage{bbding} %首先在导言区调用bbding包
\usepackage{diagbox} % 加载宏包
\usepackage{stfloats}
\usepackage{float}
\usepackage{multirow}
\usepackage{tikz}
\usetikzlibrary{arrows.meta,calc,positioning,quantikz2}

\theorembodyfont{\upshape}
\makeatletter
\renewtheoremstyle{plain}% Adds automatic line break, if heading is too long
{\item{\theorem@headerfont ##1\ ##2\theorem@separator}~}
{\item{\theorem@headerfont ##1\ ##2\ (##3)\theorem@separator}~}
\makeatother
\def\BibTeX{{\rm B\kern-.05em{\sc i\kern-.025em b}\kern-.08em
		T\kern-.1667em\lower.7ex\hbox{E}\kern-.125emX}}
\usepackage{balance}

\usepackage{enumitem}
\setenumerate[1]{itemsep=0pt,partopsep=0pt,parsep=\parskip,topsep=0pt}
\setitemize[1]{itemsep=0pt,partopsep=0pt,parsep=\parskip,topsep=0pt}
\setdescription{itemsep=0pt,partopsep=0pt,parsep=\parskip,topsep=0pt}
\begin{document}

	\title{QTrans: A Quantum Transformer for Sentiment Classification}	
	\author{Ren-Xin Zhao\textsuperscript{\textdagger},
		Xinjie Huang\textsuperscript{\textdagger},
        Yahong Liu,
		Maoyu Ye,
		Jinjing Shi\textsuperscript{*},
		Shi Wang, and Yaonan Wang%
		\IEEEcompsocitemizethanks{
            \IEEEcompsocthanksitem{This work was supported by the National Natural Science Foundation of China (Grant Nos. 62272483), the Monumental Consultation Project on the Development Strategy of Chinese Engineering and Technology (2025WK1001) and the Hunan Key Research and Development Program Project (2026QK3012).}
			\IEEEcompsocthanksitem{Ren-Xin Zhao is with the School of Computer Science, Xiangtan University, Xiangtan 411105, China (e-mail: rxz@alu.hdu.edu.cn).}
			{Xinjie Huang, Yahong Liu and Jinjing Shi are with the School of Electronic Information, Central South University, Changsha 410083, China (e-mail: huangxinjie666@outlook.com; yahongLiu2026@outlook.com; shijinjing@csu.edu.cn).}
			{Maoyu Ye is with the Department of Otorhinolaryngology-Head and Neck Surgery, the Third Xiangya Hospital, Central South University, Changsha 410013, China (e-mail: 125867490@qq.com).}
			{Shi Wang and Yaonan Wang are with the School of Artificial Intelligence and Robotics, Hunan University, Changsha 410082, China (e-mail: shi\_wang@hnu.edu.cn; yaonan@hnu.edu.cn).}
			\IEEEcompsocthanksitem{\textsuperscript{\textdagger}Ren-Xin Zhao and Xinjie Huang contributed equally to this work.}
			\IEEEcompsocthanksitem{\textsuperscript{*}Jinjing Shi is the corresponding author.}
		}
	}
	
\maketitle
	
\begin{abstract}
In small-scale binary sentiment classification scenarios, factors such as negation, contrastive shifts, and cross-word dependencies lead to the non-linear coupling of sentiment cues, making it difficult for conventional lightweight models to fully capture the contextual relationships between tokens. To address this issue, we propose a model named QTrans, which uses parameterized quantum circuits to construct query, key, and value features and derives attention coefficients from Gaussian distances between quantum measurements. By further integrating a quantum feed-forward neural network, residual connections, and layer normalization, the model establishes an end-to-end trainable quantum-classical hybrid framework for sentiment classification. Experimental results on the MR, CR, and MPQA datasets show that QTrans achieves test accuracies of 72.13\%, 69.51\%, and 63.45\%, respectively, representing improvements of 2.88, 3.17, and 3.79 percentage points over the best-performing classical baselines for each dataset. Overall, QTrans expands the application of parameterized quantum circuits in lightweight sentiment analysis and lays an experimental foundation for further research into quantum multi-head self-attention for modeling textual relationships.
\end{abstract}
	
	\begin{IEEEkeywords}
Sentiment classification, quantum Transformer, quantum multi-head self-attention, parameterized quantum circuits, quantum machine learning.
	\end{IEEEkeywords}

\section{Introduction}\label{sec:introduction}
\begingroup
\justifying%
\setlength{\emergencystretch}{2em}%

\IEEEPARstart{S}{entiment} classification has become an important means of extracting opinions from reviews, social media, and other user-generated text~\cite{1}. However, some lightweight models trained on small datasets may still rely heavily on isolated lexical cues and may underrepresent the intrinsic relationships between sentiment-bearing tokens~\cite{2,3}. As a result, a positive word may be incorrectly treated as positive evidence after being negated, the first clause may dominate the prediction even when a contrastive conjunction transfers emphasis to the second clause, and distant modifiers may be separated from the words whose meanings they change~\cite{4,5,6}. Such errors may indicate that a model has learned word-label correlations rather than the rules by which contextual cues jointly determine polarity~\cite{7}. This can reduce classification accuracy and weaken generalization in domain-specific scenarios, where additional labeled data are often difficult to obtain~\cite{8}. Expanding the model or the dataset may therefore not always be a practical remedy~\cite{9}. This motivates the development of a lightweight mechanism that can distinguish the contextual roles of tokens and model their mutual influence~\cite{10}. The self-attention mechanism provides a plausible route to this goal~\cite{11}.

Self-attention maps an input sequence into queries, keys, and values, calculates the relevance between tokens, and then aggregates the corresponding contextual information~\cite{12}. This mechanism enables a token to interact directly with both neighboring and distant tokens, making the Transformer well suited to modeling negation, contrast, and cross-word dependencies~\cite{13}. Although effective, a lightweight Transformer may remain constrained by the transformations used to establish these interactions~\cite{14}. The query, key, and value projections are typically linear mappings~\cite{16}. Once different contextual roles have been compressed into insufficiently distinguishable projected features, the subsequent attention calculation may be unable to fully recover the lost relationships~\cite{17}. Increasing the projection dimension, the number of heads, or the network depth may alleviate this limitation, but it may also introduce more parameters and a higher risk of overfitting on small datasets~\cite{18}. This trade-off motivates strengthening the representation ability of the projection process rather than relying solely on a larger classical Transformer~\cite{14,17,18}. To investigate this possibility, parameterized quantum circuits are introduced into self-attention~\cite{19}.

A parameterized quantum circuit combines data encoding, trainable quantum gates, entangling operations, and measurements to form a nonlinear feature transformation~\cite{20}. Its mechanism is structurally compatible with self-attention because independent circuits can represent the query, key, and value branches. In Gaussian-projected quantum multi-head self-attention, query and key states are reduced to expectation values whose squared distance determines the attention coefficient, while value measurements provide the contextual content to be aggregated~\cite{19}. This construction allows quantum transformations to participate in both relevance estimation and information transmission without requiring an inner product between full quantum states. The resulting contextual representation still needs a position-wise transformation and stable residual optimization~\cite{25,26,27}. Therefore, a quantum feed-forward neural network, residual connections, and layer normalization are integrated with Gaussian-projected quantum multi-head self-attention~\cite{27,28}. Constructing such a coherent quantum Transformer gives rise to the following three questions:

\begin{enumerate}
    \item Can independent quantum query, key, and value circuits be used to construct multi-head attention for modeling contextual token relationships?
    \item How can quantum multi-head self-attention and a position-wise quantum feed-forward neural network be combined with residual connections and layer normalization in a unified trainable model?
    \item Can the resulting QTrans model maintain performance advantages over lightweight classical models across different sentiment datasets?
\end{enumerate}

To this end, QTrans is proposed to answer the above questions. The main contributions of this paper are summarized as follows:

\begin{itemize}
    \item An $n$-qubit Gaussian-projected quantum multi-head self-attention mechanism is proposed. Independent parameterized quantum circuits generate query, key, and value measurements; normalized Gaussian coefficients quantify token relevance, while the measured value features are divided between two attention heads.
    \item A position-wise quantum feed-forward neural network is constructed using an independently parameterized $n$-qubit circuit. The quantum multi-head self-attention mechanism and quantum feed-forward neural network are integrated with residual connections and layer normalization to form the end-to-end trainable QTrans architecture.
    \item A four-qubit instance of QTrans is evaluated on MR \cite{29}, CR \cite{30}, and MPQA \cite{31}, achieving test accuracies of 72.13\%, 69.51\%, and 63.45\%, respectively, which exceed the strongest classical baseline on each dataset by 2.88, 3.17, and 3.79 percentage points. The ablation and embedding-noise experiments further characterize the contributions and limitations of the two quantum components.
\end{itemize}

The remainder of this paper first introduces the foundations required for the proposed model and then details the architecture and workflow of QTrans. The experimental settings and comparative results are subsequently presented and analyzed, followed by the conclusions and directions for future research.

\section{Preliminaries}\label{sec:preliminaries}

\justifying%
\setlength{\emergencystretch}{2em}%

This section provides a concise review of classical self-attention, fundamental quantum operations, and parameterized quantum circuits.

\subsection{Classical Self-Attention}\label{sec:classical_self_attention}

Let an input sequence containing $L$ elements be represented by
\begin{equation}\label{eq:self_attention_input}
X=
\begin{bmatrix}
\boldsymbol{x}_1,
\boldsymbol{x}_2,
\ldots,
\boldsymbol{x}_L
\end{bmatrix}^{\mathrm{T}}
\in\mathbb{R}^{L\times d},
\end{equation}
where $\boldsymbol{x}_i\in\mathbb{R}^{d}$ denotes the representation of the $i$-th element and $d$ is the feature dimension. The self-attention mechanism first maps $X$ into the query, key, and value matrices:
\begin{equation}\label{eq:qkv_projection}
Q=XW_Q,\qquad K=XW_K,\qquad V=XW_V,
\end{equation}
where $W_Q\in\mathbb{R}^{d\times d_k}$, $W_K\in\mathbb{R}^{d\times d_k}$, and $W_V\in\mathbb{R}^{d\times d_v}$ are learnable projection matrices. Accordingly, $Q$ describes the information requested by each element, $K$ provides the information used for matching, and $V$ contains the content to be aggregated.
The scaled dot-product attention is then expressed as \cite{13}
\begin{equation}\label{eq:scaled_dot_product_attention}
\operatorname{Attention}(Q,K,V)
=\operatorname{softmax}\!(\frac{QK^{\mathrm{T}}}{\sqrt{d_k}})V.
\end{equation}
In (\ref{eq:scaled_dot_product_attention}), $QK^{\mathrm{T}}$ measures the pairwise compatibility between queries and keys. The factor $\sqrt{d_k}$ scales the dot products, while the row-wise softmax converts them into normalized attention weights. The weighted combination of $V$ consequently produces a contextual representation for every element in the sequence.

\subsection{Fundamentals of Quantum Computing}\label{sec:quantum_computing}

The qubit is the elementary carrier of quantum information. A single-qubit pure state is written as \cite{30.1}
\begin{equation}\label{eq:qubit_state}
|\psi\rangle=\alpha|0\rangle+\beta|1\rangle,
\qquad |\alpha|^2+|\beta|^2=1,
\end{equation}
where $|0\rangle=\begin{bmatrix}1,0\end{bmatrix}^{\mathrm{T}}$ and $|1\rangle=\begin{bmatrix}0,1\end{bmatrix}^{\mathrm{T}}$ are computational basis states, and $\alpha,\beta\in\mathbb{C}$ are probability amplitudes. A general pure state of an $n$-qubit system is written as
\begin{equation}\label{eq:n_qubit_state}
|\Psi\rangle
=\sum_{b=0}^{2^n-1}c_b|b\rangle,
\qquad
\sum_{b=0}^{2^n-1}|c_b|^2=1.
\end{equation}
A state is separable if it can be factorized as $|\Psi\rangle=\otimes_{j=1}^{n}|\psi_j\rangle$; otherwise, it is entangled and cannot be represented as a product of individual qubit states.
Quantum states evolve through unitary gates. A rotation about the $\mu$ axis, with $\mu\in\{x,y\}$, is generally defined by
\begin{equation}\label{eq:rotation_gate}
R_{\mu}(\vartheta)
=\exp\!(-\frac{\mathrm{i}\vartheta}{2}\sigma_{\mu}),
\end{equation}
where $\vartheta$ is the rotation angle and $\sigma_{\mu}$ is the corresponding Pauli matrix. Multi-qubit correlations can be introduced by entangling gates. For example, the controlled-NOT gate acts on computational basis states according to
\begin{equation}\label{eq:cnot_gate}
\operatorname{CNOT}|a,b\rangle=|a,a\oplus b\rangle,
\qquad a,b\in\{0,1\},
\end{equation}
where $\oplus$ denotes addition modulo two. After the quantum state has been prepared and transformed, information can be extracted through measurement. For a Hermitian observable $P$, its expectation value is
\begin{equation}\label{eq:expectation_value}
\langle P\rangle_{\psi}=\langle\psi|P|\psi\rangle,
\end{equation}
which is a real-valued quantity determined by the measured quantum state and the selected observable.

The quantum gates and measurement observable used in QTrans are summarized in Tab.~\ref{tab:qtrans_quantum_gates}. The Pauli-$Z$ operator is measured at the circuit output, whereas the remaining entries are used for state preparation, data encoding, trainable rotations, or qubit entanglement. The CNOT matrix is written in the ordered computational basis $\{|00\rangle,|01\rangle,|10\rangle,|11\rangle\}$, with the first qubit acting as the control.

\begin{table*}[t]
\centering
\caption{Quantum gates and measurement observable used in QTrans.}
\label{tab:qtrans_quantum_gates}
\small
\renewcommand{\arraystretch}{1.65}
\setlength{\tabcolsep}{10pt}
\begin{tabular}{@{}l c c c@{}}
\toprule
Gate or observable & Mathematical symbol & Matrix representation & Circuit symbol \\
\midrule
Hadamard gate
& $H$
& $\displaystyle \frac{1}{\sqrt{2}}\begin{bmatrix}1&1\\1&-1\end{bmatrix}$
& \begin{quantikz}[row sep=0.05cm,column sep=0.10cm,scale=0.80,transform shape]
    \qw & \gate[style={qgate h}]{H} & \qw
  \end{quantikz} \\
%Pauli-$Z$ observable
%& $Z$
%& $\displaystyle \begin{bmatrix}1&0\\0&-1\end{bmatrix}$
%& \begin{quantikz}[row sep=0.05cm,column sep=0.10cm,scale=0.80,transform shape]
 %   \qw & \meter{\ensuremath{Z}}
 % \end{quantikz} \\
$X$  gate
& $R_X(\vartheta)$
& $\displaystyle \begin{bmatrix}
\cos\!(\frac{\vartheta}{2})&-\mathrm{i}\sin\!(\frac{\vartheta}{2})\\
-\mathrm{i}\sin\!(\frac{\vartheta}{2})&\cos\!(\frac{\vartheta}{2})
\end{bmatrix}$
& \begin{quantikz}[row sep=0.05cm,column sep=0.10cm,scale=0.80,transform shape]
    \qw & \gate[style={qgate rx}]{\ensuremath{R_X(\vartheta)}} & \qw
  \end{quantikz} \\
$Y$  gate
& $R_Y(\vartheta)$
& $\displaystyle \begin{bmatrix}
\cos\!(\frac{\vartheta}{2})&-\sin\!(\frac{\vartheta}{2})\\
\sin\!(\frac{\vartheta}{2})&\cos\!(\frac{\vartheta}{2})
\end{bmatrix}$
& \begin{quantikz}[row sep=0.05cm,column sep=0.10cm,scale=0.80,transform shape]
    \qw & \gate[style={qgate ry}]{\ensuremath{R_Y(\vartheta)}} & \qw
  \end{quantikz} \\
Controlled-NOT gate
& $\operatorname{CNOT}$
& $\displaystyle \begin{bmatrix}
1&0&0&0\\0&1&0&0\\0&0&0&1\\0&0&1&0
\end{bmatrix}$
& \begin{quantikz}[row sep=0.08cm,column sep=0.10cm,scale=0.80,transform shape]
    \qw & \ctrl[style={qgate cnot},wire style={qwire cnot}]{1} & \qw\\
    \qw & \targ[style={qgate cnot}]{} & \qw
  \end{quantikz} \\
\bottomrule
\end{tabular}
\end{table*}

\subsection{Parameterized Quantum Circuits}\label{sec:parameterized_quantum_circuits}

A parameterized quantum circuit is a trainable unitary process whose behavior is controlled by classical parameters. Given an input vector $\boldsymbol{x}$ and a parameter vector $\boldsymbol{\theta}$, a general circuit prepares the state \cite{30.1}
\begin{equation}\label{eq:pqc_state}
|\phi(\boldsymbol{x},\boldsymbol{\theta})\rangle
=U_{\mathrm{var}}(\boldsymbol{\theta})
U_{\mathrm{enc}}(\boldsymbol{x})
|0\rangle^{\otimes n},
\end{equation}
where $U_{\mathrm{enc}}(\boldsymbol{x})$ is an input-dependent unitary operator and $U_{\mathrm{var}}(\boldsymbol{\theta})$ is a trainable unitary operator. The circuit output associated with an observable $P$ is defined as
\begin{equation}\label{eq:pqc_output}
f_{\boldsymbol{\theta}}(\boldsymbol{x})
=\langle\phi(\boldsymbol{x},\boldsymbol{\theta})|
P
|\phi(\boldsymbol{x},\boldsymbol{\theta})\rangle.
\end{equation}
Equation (\ref{eq:pqc_output}) defines a parameter-dependent mapping from a classical input to a real-valued output. The mapping can be adapted to a learning objective by optimizing $\boldsymbol{\theta}$. Its functional form is jointly determined by the encoding operator, the variational unitary, and the measured observable.

\section{Quantum Transformer for Sentiment Classification}\label{sec:qtrans}

In this section, QTrans is constructed for binary sentiment classification. As shown in Fig.~\ref{fig:qtrans_framework}, it contains four successive stages: sequence embedding, quantum multi-head self-attention, a quantum feed-forward neural network, and sentiment readout \cite{37}. Gaussian similarities between measured quantum query and key features are employed to quantify token relevance \cite{19}. The overall framework and the two quantum components are introduced in the following subsections.

\subsection{Overall Framework}\label{sec:qtrans_framework}

The overall framework of QTrans is shown in Fig.~\ref{fig:qtrans_framework}. Given a sentence, the tokenizer inserts a start token $[\mathrm{CLS}]$ and an end token $[\mathrm{SEP}]$, after which the sequence is truncated or padded to a fixed length $L$. Let
\begin{equation}\label{eq:qtrans_token_sequence}
S=[t_1,t_2,\ldots,t_L],
\qquad t_1=[\mathrm{CLS}],
\end{equation}
where the last valid element is $[\mathrm{SEP}]$. Token and positional embeddings are added to form
\begin{equation}\label{eq:qtrans_input_embedding}
\boldsymbol{x}_i
=E_T(t_i)+E_P(i),
X=[\boldsymbol{x}_1,\ldots,\boldsymbol{x}_L]^{\mathrm T}
\in\mathbb{R}^{L\times n}.
\end{equation}
Here, $E_T$ and $E_P$ denote learnable token and positional embeddings, respectively. The embedding width $n$ is matched to the number of qubits in each quantum circuit; the experimental model uses $n=4$. The sentiment-label row in Fig.~\ref{fig:qtrans_framework} denotes the supervision associated with the complete sequence rather than token-level labels. A single target $y\in\{0,1\}$ is used in the loss function and is not included in $X$.

Let $m_i=1$ for a valid token and $m_i=0$ for padding. The lower-triangular causal-padding mask used by the attention module is
\begin{equation}\label{eq:qtrans_attention_mask}
B_{ij}=
\begin{cases}
1, & m_i=m_j=1\ \text{and}\ j\leq i,\\
0, & \text{otherwise}.
\end{cases}
\end{equation}
Thus, a valid position attends only to itself and earlier valid positions. In particular, the terminal $[\mathrm{SEP}]$ position can aggregate information from the complete sentence and is consequently used as the classification readout.

\begin{figure*}[!htbp]
\centering
\includegraphics[width=1\textwidth]{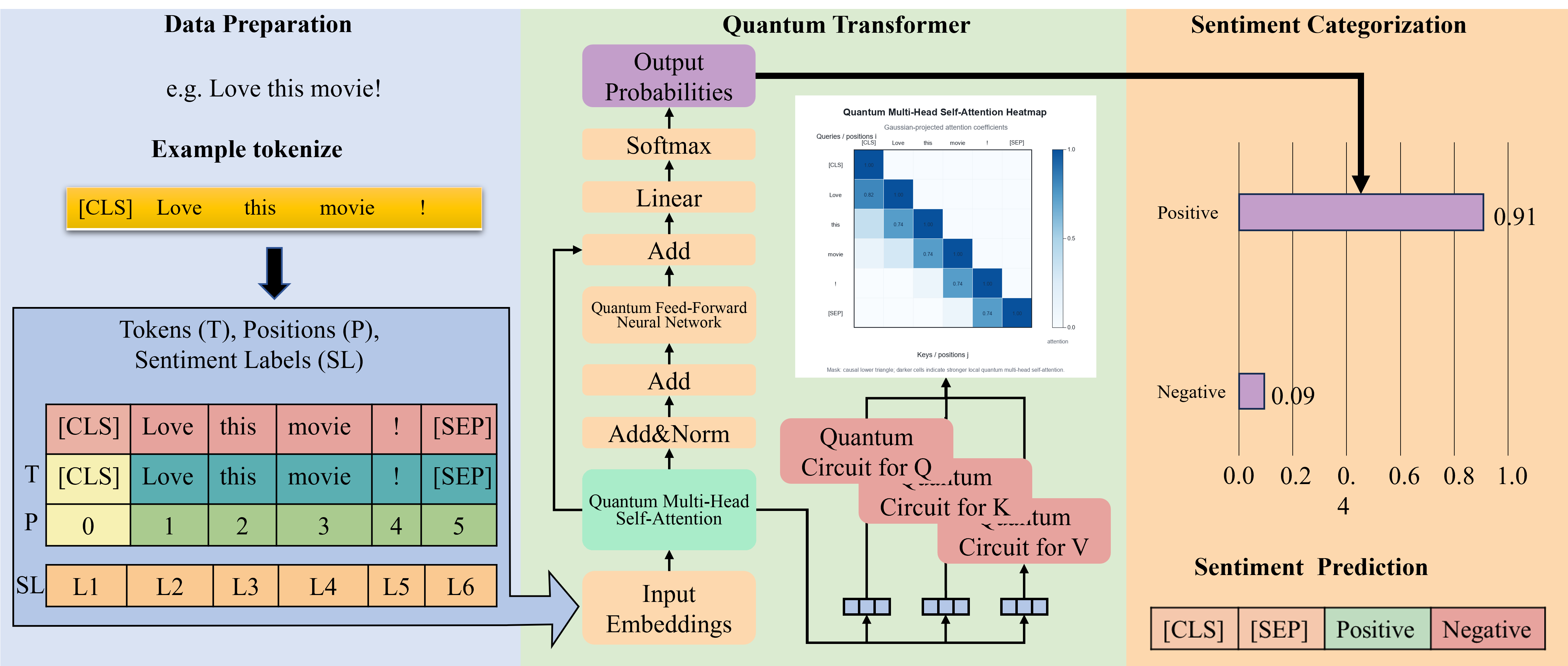}
\caption{The overall framework of QTrans. A sentence is converted into $[\mathrm{CLS}]$-to-$[\mathrm{SEP}]$ tokens and positional indices; the sentiment label is retained only as the training target. The input embeddings are processed by Gaussian-projected quantum multi-head self-attention constructed from quantum query, key, and value circuits. The heatmap schematically displays the masked Gaussian coefficients before row normalization. Residual addition and layer normalization are followed by a quantum feed-forward neural network and a second residual addition. The representation at the terminal $[\mathrm{SEP}]$ position is mapped to positive and negative logits, and softmax converts them into output probabilities.}
\label{fig:qtrans_framework}
\end{figure*}

Denote the output of quantum multi-head self-attention by $A\in\mathbb{R}^{L\times n}$. The first residual and normalization operation in Fig.~\ref{fig:qtrans_framework} is
\begin{equation}\label{eq:qtrans_first_residual}
H_1=\operatorname{LN}
(X+\operatorname{Dropout}(A)).
\end{equation}
The residual path preserves the token and positional information, while $A$ introduces context gathered according to the quantum multi-head self-attention coefficients. The quantum feed-forward neural network then produces $F\in\mathbb{R}^{L\times n}$, and the second Add node gives
\begin{equation}\label{eq:qtrans_second_residual}
H_2=H_1+\operatorname{Dropout}(F).
\end{equation}
Layer normalization is applied after the first residual addition. The representation $H_1$ is subsequently transmitted to the quantum feed-forward neural network and retained on the second residual path. The output of the quantum feed-forward neural network is merged with $H_1$ by the Add operation in \eqref{eq:qtrans_second_residual}.

Let $i_{\mathrm{sep}}$ be the index of the last valid token. The sentence representation and binary logits are
\begin{equation}\label{eq:qtrans_classifier}
\begin{aligned}
\boldsymbol{z}&=H_2[i_{\mathrm{sep}},:]^{\mathrm T}\in\mathbb{R}^{n},\\
\boldsymbol{o}
&=W_c^{\mathrm T}\operatorname{Dropout}(\boldsymbol{z})+\boldsymbol{b}_c,
\quad
W_c\in\mathbb{R}^{n\times2},
\quad
\boldsymbol{b}_c\in\mathbb{R}^{2}.
\end{aligned}
\end{equation}
The probability vector displayed in Fig.~\ref{fig:qtrans_framework} is $\boldsymbol{p}=\operatorname{softmax}(\boldsymbol{o})$, whose two components represent the positive and negative sentiment probabilities according to the label order fixed during preprocessing. For a mini-batch containing $N$ samples, the label-smoothed cross-entropy used in Section~\ref{sec:experimental_settings} is
\begin{equation}\label{eq:qtrans_training_loss}
\begin{aligned}
\widetilde{y}_{b,c}
&=(1-\epsilon)\mathbb{I}(c=y_b)+\frac{\epsilon}{2},\\
\mathcal{L}
&=-\frac{1}{N}\sum_{b=1}^{N}\sum_{c=0}^{1}
\widetilde{y}_{b,c}
\log\frac{\exp(o_{b,c})}
{\sum_{r=0}^{1}\exp(o_{b,r})}.
\end{aligned}
\end{equation}
During training, the loss is evaluated directly from the logits for numerical stability; softmax is explicitly applied when class probabilities are required for inference.

\subsection{Quantum Multi-Head Self-Attention}\label{sec:quantum_multihead_attention}

The quantum multi-head self-attention module replaces the scaled dot product in \eqref{eq:scaled_dot_product_attention} with a Gaussian similarity computed from measured quantum features. Three independently parameterized circuits generate the query, key, and value features of each token. The measured query-key distance determines token relevance, while the measured value features provide the contextual information to be aggregated \cite{19}.

The circuit width $n$ is taken to be even with $n\geq4$, allowing the measured value vector to be divided evenly between two attention heads. For $\boldsymbol{x}_i=[x_{i,1},\ldots,x_{i,n}]^{\mathrm T}$, the embedding components are bounded and encoded as rotation angles:
\begin{equation}\label{eq:qtrans_angle_encoding}
\phi_{i,j}=\pi\tanh(x_{i,j}),
\qquad j=1,\ldots,n.
\end{equation}
The $n$-qubit encoding operator is
\begin{equation}\label{eq:qtrans_encoding_state}
\mathcal{E}(\boldsymbol{x}_i)
=(\otimes_{j=1}^{n}R_Y(\phi_{i,j}))H^{\otimes n}.
\end{equation}
The initial Hadamard layer creates a superposition, and the data-dependent rotations inject the token representation. The branch state is obtained from the general circuit definition in \eqref{eq:pqc_state} by setting $U_{\mathrm{enc}}=\mathcal{E}$ and $U_{\mathrm{var}}=U_r$.

For each branch $r\in\{Q,K,V,F\}$, QTrans uses an independent strongly entangling ansatz. Its initial trainable layer is
\begin{equation}\label{eq:qtrans_initial_rotation}
U_{r,0}
=\otimes_{j=1}^{n}
R_Y(\theta^{r,y}_{0,j})R_X(\theta^{r,x}_{0,j}),
\end{equation}
so that $R_X$ is applied before $R_Y$. The ring entangler and the $\ell$-th repeated block are
\begin{equation}\label{eq:qtrans_entangling_ansatz}
\begin{aligned}
U_{\mathrm{ring}}^{(n)}
&=\operatorname{CNOT}_{narrow1}
\operatorname{CNOT}_{n-1arrow n}\\
&\quad\cdots
\operatorname{CNOT}_{2arrow3}
\operatorname{CNOT}_{1arrow2},\\
U_{r,\ell}
&=(\otimes_{j=1}^{n}R_Y(\theta^r_{\ell,j}))
U_{\mathrm{ring}}^{(n)},
\qquad \ell=1,\ldots,D_q,\\
U_r(\boldsymbol{\theta}_r)
&=U_{r,D_q}\cdots U_{r,1}U_{r,0}.
\end{aligned}
\end{equation}
Accordingly, every repeated block first introduces ring entanglement and then applies trainable $R_Y$ rotations. The complete $n$-qubit circuit specified by \eqref{eq:qtrans_encoding_state}, \eqref{eq:qtrans_initial_rotation}, and \eqref{eq:qtrans_entangling_ansatz} is drawn in Fig.~\ref{fig:qtrans_quantum_circuit}. The experiments instantiate it with $n=4$ and $D_q=2$.

\begin{figure*}[!htbp]
\centering
\begin{quantikz}[row sep=0.22cm, column sep=0.18cm, scale=0.80, transform shape]
\lstick{\ensuremath{\lvert0\rangle_1}}
& \gate[style={qgate h}]{H}
& \gate[style={qgate ry}]{\ensuremath{R_Y(\phi_{i,1})}}
& \gate[style={qgate rx}]{\ensuremath{R_X(\theta^{r,x}_{0,1})}}
& \gate[style={qgate ry}]{\ensuremath{R_Y(\theta^{r,y}_{0,1})}}
& \gate[wires=4,style={draw={rgb,255:red,208;green,105;blue,41},fill={rgb,255:red,255;green,242;blue,226},text={rgb,255:red,156;green,71;blue,26},rounded corners=2pt,line width=0.6pt}]{\ensuremath{U_{\mathrm{ring}}^{(n)}}}\gategroup[
	wires=4,
	steps=2,
	style={dashed,rounded corners,fill=green!4,inner xsep=3pt},
	background,
	label style={label position=below,anchor=north,yshift=-0.10cm}
]{\ensuremath{\times D_q}}
& \gate[style={qgate ry}]{\ensuremath{R_Y(\theta^r_{\ell,1})}}
& \meter[style={qgate measure}]{\ensuremath{Z_1}} \\
\lstick{\ensuremath{\lvert0\rangle_2}}
& \gate[style={qgate h}]{H}
& \gate[style={qgate ry}]{\ensuremath{R_Y(\phi_{i,2})}}
& \gate[style={qgate rx}]{\ensuremath{R_X(\theta^{r,x}_{0,2})}}
& \gate[style={qgate ry}]{\ensuremath{R_Y(\theta^{r,y}_{0,2})}}
&
& \gate[style={qgate ry}]{\ensuremath{R_Y(\theta^r_{\ell,2})}}
& \meter[style={qgate measure}]{\ensuremath{Z_2}} \\
\lstick{\ensuremath{\lvert0\rangle_j,\;3\leq j\leq n-1}}
& \gate[style={qgate h}]{H}
& \gate[style={qgate ry}]{\ensuremath{R_Y(\phi_{i,j})}}
& \gate[style={qgate rx}]{\ensuremath{R_X(\theta^{r,x}_{0,j})}}
& \gate[style={qgate ry}]{\ensuremath{R_Y(\theta^{r,y}_{0,j})}}
&
& \gate[style={qgate ry}]{\ensuremath{R_Y(\theta^r_{\ell,j})}}
& \meter[style={qgate measure}]{\ensuremath{Z_j}} \\
\lstick{\ensuremath{\lvert0\rangle_n}}
& \gate[style={qgate h}]{H}
& \gate[style={qgate ry}]{\ensuremath{R_Y(\phi_{i,n})}}
& \gate[style={qgate rx}]{\ensuremath{R_X(\theta^{r,x}_{0,n})}}
& \gate[style={qgate ry}]{\ensuremath{R_Y(\theta^{r,y}_{0,n})}}
&
& \gate[style={qgate ry}]{\ensuremath{R_Y(\theta^r_{\ell,n})}}
& \meter[style={qgate measure}]{\ensuremath{Z_n}}
\end{quantikz}%
\caption{The $n$-qubit parameterized quantum circuit used by branch $r\in\{Q,K,V,F\}$. The intermediate wire indexed by $j$ represents every qubit from 3 to $n-1$. A Hadamard layer and data-dependent $R_Y$ rotations encode the token representation. The trainable ansatz begins with an $R_X$-$R_Y$ layer and then repeats the $n$-qubit CNOT ring $U_{\mathrm{ring}}^{(n)}$ followed by trainable $R_Y$ rotations $D_q$ times. The query and key branches retain the $Z_1$ and $Z_2$ measurements for the two attention heads, whereas the value and feed-forward branches retain all $n$ measurements.}
\label{fig:qtrans_quantum_circuit}
\end{figure*}
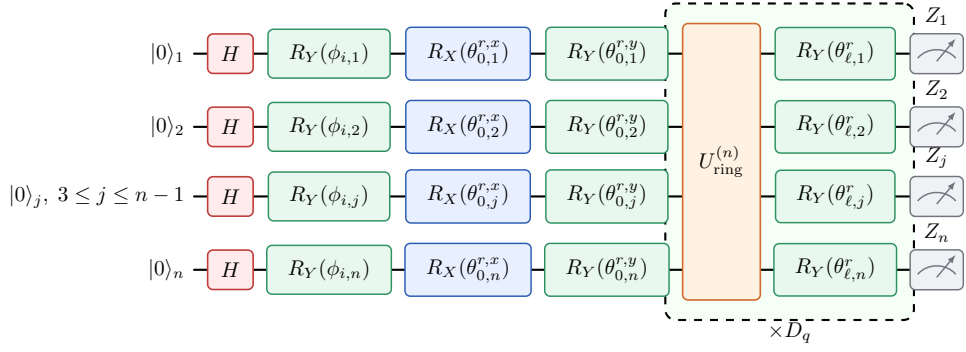

As shown in Fig.~\ref{fig:qtrans_quantum_circuit}, the same $n$-qubit circuit topology is employed by the four branches, while their trainable parameters remain independent. The query and key branches use the first two Pauli-$Z$ measurements to form the two attention heads. The value branch and the quantum feed-forward neural network branch use all $n$ measurements so that their outputs retain the $n$-dimensional residual width. Thus, Fig.~\ref{fig:qtrans_quantum_circuit} gives the common circuit realization of the quantum multi-head self-attention mechanism and the quantum feed-forward neural network.

For branch $r$ and observable $Z_j$, denote the output of \eqref{eq:pqc_output} by
\begin{equation}\label{eq:qtrans_branch_measurement}
z_{i,j}^{(r)}
=f_{\boldsymbol{\theta}_r}^{[Z_j]}(\boldsymbol{x}_i),
\qquad r\in\{Q,K,V\},\quad j=1,\ldots,n,
\end{equation}
where $Z_j$ denotes the Pauli-$Z$ operator on the $j$-th qubit and the identity operator on the other qubits. The superscript $[Z_j]$ identifies the observable in \eqref{eq:pqc_output}, while the encoding and variational operators are given by \eqref{eq:qtrans_encoding_state} and \eqref{eq:qtrans_entangling_ansatz}, respectively. QTrans forms two measurement heads without duplicating circuit parameters. For $h\in\{1,2\}$, the query and key scalars are
\begin{equation}\label{eq:qtrans_quantum_query_key}
\begin{aligned}
q_i^{(h)}&=z_{i,h}^{(Q)},\\
k_i^{(h)}&=z_{i,h}^{(K)}.
\end{aligned}
\end{equation}
The value circuit is measured on all $n$ qubits:
\begin{equation}\label{eq:qtrans_quantum_value}
\boldsymbol{v}_i
=\begin{bmatrix}
z_{i,1}^{(V)} & z_{i,2}^{(V)} & \cdots & z_{i,n}^{(V)}
\end{bmatrix}^{\mathrm T}.
\end{equation}
For the two-head configuration, $n$ is even and the value vector is divided into two equal parts: $\boldsymbol{v}_i^{(1)}=[v_{i,1},\ldots,v_{i,n/2}]^{\mathrm T}$ and $\boldsymbol{v}_i^{(2)}=[v_{i,n/2+1},\ldots,v_{i,n}]^{\mathrm T}$. Thus, each head has a scalar quantum query, a scalar quantum key, and an $n/2$-dimensional quantum value.

For an admissible query-key pair, the unnormalized Gaussian coefficient is
\begin{equation}\label{eq:qtrans_gaussian_attention}
g_{ij}^{(h)}
=B_{ij}\exp\![-(q_i^{(h)}-k_j^{(h)})^2].
\end{equation}
The lower-triangular heatmap in Fig.~\ref{fig:qtrans_framework} illustrates the masked coefficients $g_{ij}^{(h)}$ for one attention head. These coefficients are normalized before value aggregation.
For every valid query position, normalization over the admissible keys gives
\begin{equation}\label{eq:qtrans_normalized_attention}
\alpha_{ij}^{(h)}
=\begin{cases}
\displaystyle\frac{g_{ij}^{(h)}}
{\sum_{u=1}^{L}g_{iu}^{(h)}}, & m_i=1,\\[6pt]
0, & m_i=0,
\end{cases}
\qquad
\boldsymbol{c}_i^{(h)}
=\sum_{j=1}^{L}\alpha_{ij}^{(h)}\boldsymbol{v}_j^{(h)}.
\end{equation}
Because $B_{ij}=0$ for future and padded positions, these positions receive exactly zero attention weight. Equations~\eqref{eq:qtrans_gaussian_attention} and \eqref{eq:qtrans_normalized_attention} are equivalent to applying a masked softmax to the negative squared distance between the measured query and key features. The normalized coefficients $\alpha_{ij}^{(h)}$ are the actual weights used to aggregate the quantum value features.

The two head outputs are concatenated and fused within the attention block:
\begin{equation}\label{eq:qtrans_attention_output}
\begin{aligned}
C[i,:]
&=\operatorname{Concat}\!(
(\boldsymbol{c}_i^{(1)})^{\mathrm T},
(\boldsymbol{c}_i^{(2)})^{\mathrm T}),\\
A&=CW_O,
\qquad W_O\in\mathbb{R}^{n\times n}.
\end{aligned}
\end{equation}
Here, $C,A\in\mathbb{R}^{L\times n}$. The quantum circuits determine both the attention coefficients and the content being aggregated, whereas the small classical matrix $W_O$ mixes the two measurement heads and preserves the residual width. The residual connection is then applied once, outside the attention calculation, through \eqref{eq:qtrans_first_residual}.

\subsection{Quantum Feed-Forward Neural Network}\label{sec:quantum_feedforward}

The quantum feed-forward neural network transforms each token representation independently after quantum multi-head self-attention has established contextual dependencies. Its parameter set $\boldsymbol{\theta}_F$ is independent of $\boldsymbol{\theta}_Q$, $\boldsymbol{\theta}_K$, and $\boldsymbol{\theta}_V$. For the quantum feed-forward neural network branch, the input is encoded by \eqref{eq:qtrans_encoding_state} and transformed by the circuit in Fig.~\ref{fig:qtrans_quantum_circuit}, which is mathematically specified by \eqref{eq:qtrans_initial_rotation} and \eqref{eq:qtrans_entangling_ansatz} with $r=F$.

For the $i$-th row $\boldsymbol{h}_{1,i}$ of $H_1$, define
\begin{equation}\label{eq:qtrans_quantum_feedforward_projection}
G_{i,j}
=f_{\boldsymbol{\theta}_F}^{[Z_j]}(\boldsymbol{h}_{1,i}),
\qquad j=1,\ldots,n,
\end{equation}
where the quantum output is evaluated according to \eqref{eq:pqc_output} after replacing $\boldsymbol{x}_i$ with $\boldsymbol{h}_{1,i}$ in \eqref{eq:qtrans_encoding_state}. Collecting the measurements gives $G\in\mathbb{R}^{L\times n}$. The position-wise feed-forward output is
\begin{equation}\label{eq:qtrans_quantum_feedforward_definition}
F=\operatorname{GELU}(G)W_F
+\boldsymbol{1}_L\boldsymbol{b}_F^{\mathrm T},
\quad
W_F\in\mathbb{R}^{n\times n},
\quad
\boldsymbol{b}_F\in\mathbb{R}^{n}.
\end{equation}
The quantum projection supplies a nonlinear, entanglement-dependent feature map, GELU introduces a classical activation, and the final linear mapping returns the features to the $n$-dimensional residual space. Since this transformation is applied row by row, it refines the context at each token without constructing a second token-to-token attention matrix. Its output enters the residual addition in \eqref{eq:qtrans_second_residual} before the terminal $[\mathrm{SEP}]$ representation is classified.

For $n$ qubits and a repeated depth $D_q$, each trainable ansatz contains $2n$ parameters in the initial $R_X$-$R_Y$ layer and $nD_q$ parameters in the repeated entangling blocks, giving $n(D_q+2)$ quantum parameters. The query, key, value, and feed-forward circuits therefore contain $4n(D_q+2)$ quantum parameters in total. With $n=4$ and $D_q=2$, this gives 16 parameters per circuit and 64 quantum parameters for the four branches. The two attention heads are obtained from different measurements and do not introduce additional circuit parameters. All quantum parameters are jointly optimized with the embeddings, classical head-fusion and feed-forward matrices, normalization layer, and classifier by backpropagating the loss in \eqref{eq:qtrans_training_loss} through the complete hybrid model.

\section{Experiments}\label{sec:experiments}

In this section, QTrans is implemented with PyTorch and DeepQuantum \cite{32} to perform binary sentiment classification. Specifically, the experiments are divided into the following three parts:

\begin{itemize}
    \item The quantum multi-head self-attention mechanism and quantum feed-forward neural network are separately replaced by their classical counterparts to clarify the contribution of the two quantum components in QTrans.
    \item QTrans is compared with Tiny Transformer \cite{33}, BiLSTM-Attention \cite{34}, and TextCNN \cite{35} under identical data partitions and nearly equal parameter scales to evaluate its classification performance.
    \item Gaussian noise with standard deviations of $0.1$, $0.3$, and $0.5$ is injected into the token embeddings to examine the sensitivity of QTrans to input perturbations.
\end{itemize}

\subsection{Datasets}\label{sec:datasets}

MR is a sentence-level movie-review polarity dataset designed to distinguish positive and negative opinions. CR is collected from customer reviews and reflects users' evaluations of product attributes and overall experience. MPQA contains opinion expressions with contextual polarity annotations and is widely used for polarity recognition. Together, these datasets cover movie reviews, product reviews, and general opinion expressions, thereby providing three sentiment classification scenarios with different linguistic characteristics.

Duplicate texts are first removed from each dataset. The remaining samples are stratified into training, validation, and test sets at a ratio of $70\%:15\%:15\%$. To prevent class priors from influencing accuracy, the positive and negative samples within every split are independently downsampled to a strict $1:1$ ratio. This procedure also ensures that no sample crosses the boundary between training, validation, and testing. The resulting dataset sizes are listed in Tab.~\ref{tab:dataset_statistics}.

\begin{table}[!th]
\centering
\caption{Dataset statistics after strict class balancing.}
\label{tab:dataset_statistics}
\begin{tabular}{lrrrr}
\toprule
Dataset & Training & Validation & Test & Total \\
\midrule
MR   & 7462 & 1598 & 1600 & 10660 \\
CR   & 1910 &  410 &  410 &  2730 \\
MPQA & 3334 &  716 &  714 &  4764 \\
\midrule
Total & 12706 & 2724 & 2724 & 18154 \\
\bottomrule
\end{tabular}
\end{table}

\subsection{Experimental Settings}\label{sec:experimental_settings}

Tab.~\ref{tab:main_configuration} delineates the principal experimental configuration. The generic $n$-qubit circuit is instantiated with $n=4$, a repeated entangling depth of $D_q=2$, and two attention heads. The embedding table reserves 6000 base-vocabulary entries, to which $[\mathrm{CLS}]$ and $[\mathrm{SEP}]$ are added as two sequence markers. Each sequence is truncated or padded to 64 tokens, and every token is represented by a four-dimensional embedding. The model is trained for 120 epochs with AdamW~\cite{36}, label-smoothed cross-entropy, gradient clipping, and cosine learning-rate annealing.

\begin{table}[!htbp]
\centering
\caption{Principal experimental configuration.}
\label{tab:main_configuration}
\begin{tabular}{ll}
\toprule
Indicator & Configuration \\
\midrule
Quantum platform & DeepQuantum \\
Classical framework & PyTorch \\
Number of qubits & 4 \\
Repeated circuit depth $D_q$ & 2 \\
Reserved base vocabulary & 6000 (+2 markers) \\
Maximum sequence length & 64 \\
Embedding dimension & 4 \\
Number of attention heads & 2 \\
Batch size & 64 \\
Training epochs & 120 \\
Optimizer & AdamW \\
Initial learning rate & $5\times10^{-4}$ \\
Weight decay & $3\times10^{-2}$ \\
Dropout & 0.25 \\
Label smoothing & 0.08 \\
Token dropout & 0.05 \\
Gradient clipping & 1.0 \\
\bottomrule
\end{tabular}
\end{table}

All compared models share the same vocabulary, sequence length, data partitions, batch size, number of epochs, optimizer, label smoothing, and token-dropout strategy. Every epoch produces genuine training, validation, and test measurements. An exponential moving average with a window of 15 epochs is used only to reduce visual fluctuation in the learning curves; the underlying records are neither replaced by cumulative optima nor truncated or manually modified. The checkpoint with the lowest validation cross-entropy is used for the final comparison, and the test set does not participate in hyperparameter selection.

To further ensure a fair comparison, the three classical baselines also use four-dimensional token embeddings, and their hidden dimensions are reduced accordingly. The QTrans total in Tab.~\ref{tab:parameter_scale} consists of $(6000+2)\times4$ token-embedding parameters, $64\times4$ positional-embedding parameters, 64 quantum parameters, 16 attention-fusion parameters, 20 feed-forward parameters, 8 layer-normalization parameters, and 10 classifier parameters, giving 24382 trainable parameters. The parameter difference between QTrans and every baseline is below $1\%$. Therefore, the following performance comparison is not established by granting QTrans a substantially larger model capacity.

\begin{table}[!htbp]
\centering
\caption{Trainable parameter scales of the four models.}
\label{tab:parameter_scale}
\begin{tabular}{lrr}
\toprule
Model & Parameters & Relative difference \\
\midrule
QTrans & 24382 & 0 \\
Tiny Transformer & 24518 & 0.56\% \\
BiLSTM-Attention & 24151 & 0.95\% \\
TextCNN & 24140 & 0.99\% \\
\bottomrule
\end{tabular}
\end{table}

\subsection{Experimental Analysis}\label{sec:experimental_analysis}

\subsubsection{Ablation Analysis}\label{sec:ablation_analysis}

The ablation experiment contains two controlled comparisons, as shown in Fig.~\ref{fig:ablation_experiment}. The left panel reports the results obtained by replacing classical self-attention with quantum multi-head self-attention on MR, CR, and MPQA. The right panel reports the results obtained by replacing the classical feed-forward module with the quantum feed-forward neural network on the same datasets. In each comparison, the remaining model components and experimental settings are kept unchanged. All models are trained for 120 epochs, and the bars show the test accuracy obtained from the checkpoint selected by the lowest validation cross-entropy. The ablation study and the model comparison in Fig.~\ref{fig:model_comparison_experiment} are trained independently; therefore, the absolute checkpoint values are used only for comparisons within their corresponding experiments.

\begin{figure*}[!htbp]
\centering
\includegraphics[width=\textwidth]{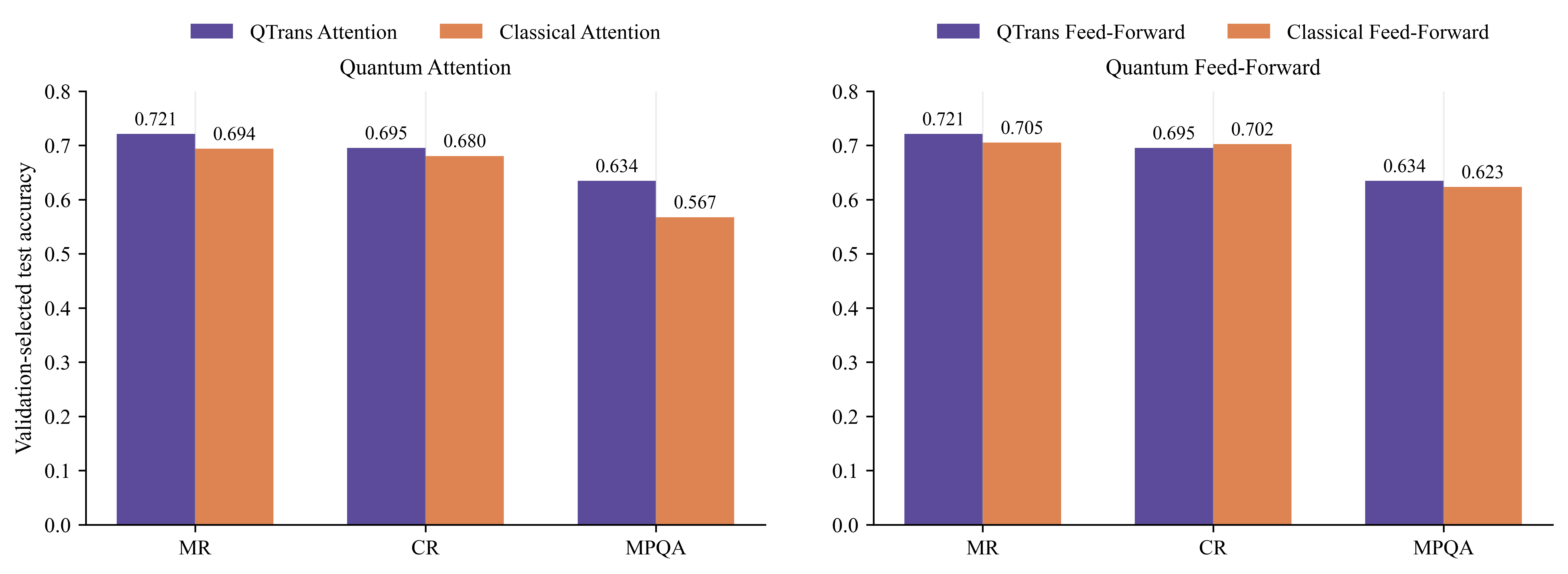}
\caption{Experiment 1: component-wise ablation results on MR, CR, and MPQA. The left panel shows the effect of replacing classical self-attention with quantum multi-head self-attention, while the right panel shows the effect of replacing the classical feed-forward module with the quantum feed-forward neural network. All bars report validation-selected test accuracy.}
\label{fig:ablation_experiment}
\end{figure*}

According to Fig.~\ref{fig:ablation_experiment}, the following conclusions can be drawn.

\begin{itemize}
    \item After classical self-attention is replaced by quantum multi-head self-attention, the test accuracy increases from $69.4\%$ to $72.1\%$ on MR, from $68.0\%$ to $69.5\%$ on CR, and from $56.7\%$ to $63.4\%$ on MPQA. The corresponding improvements are 2.7, 1.5, and 6.7 percentage points, respectively. Thus, quantum multi-head self-attention achieves higher accuracy than classical self-attention on all three datasets.
    \item After the classical feed-forward module is replaced by the quantum feed-forward neural network, the test accuracy changes from $70.5\%$ to $72.1\%$ on MR, from $70.2\%$ to $69.5\%$ on CR, and from $62.3\%$ to $63.4\%$ on MPQA. The quantum feed-forward neural network therefore improves the accuracy by 1.6 and 1.1 percentage points on MR and MPQA, respectively, but decreases it by 0.7 percentage points on CR.
    \item Compared with the feed-forward replacement, the self-attention replacement produces a more consistent improvement across datasets. In particular, the 6.7-percentage-point increase on MPQA indicates that the quantum query, key, and value projections contribute substantially to contextual polarity modeling on this dataset.
\end{itemize}

Overall, Fig.~\ref{fig:ablation_experiment} demonstrates that replacing classical self-attention with quantum multi-head self-attention improves performance consistently across the three datasets, whereas replacing the classical feed-forward module with the quantum feed-forward neural network yields dataset-dependent results.

\subsubsection{Performance Beyond Lightweight Classical Models}\label{sec:model_comparison}

QTrans is subsequently compared with three lightweight classical models. Tiny Transformer represents a compact classical self-attention architecture with the same embedding dimension; BiLSTM-Attention combines bidirectional recurrent encoding with attention-based pooling; and TextCNN employs one-dimensional convolutional kernels of widths 3, 5, and 7 to extract local patterns. Fig.~\ref{fig:model_comparison_experiment} presents the complete training dynamics, while Tab.~\ref{tab:model_comparison_results} reports the validation-selected test accuracy and macro-averaged F1 score of all four models.

\begin{figure*}[!htbp]
\centering
\includegraphics[width=\textwidth]{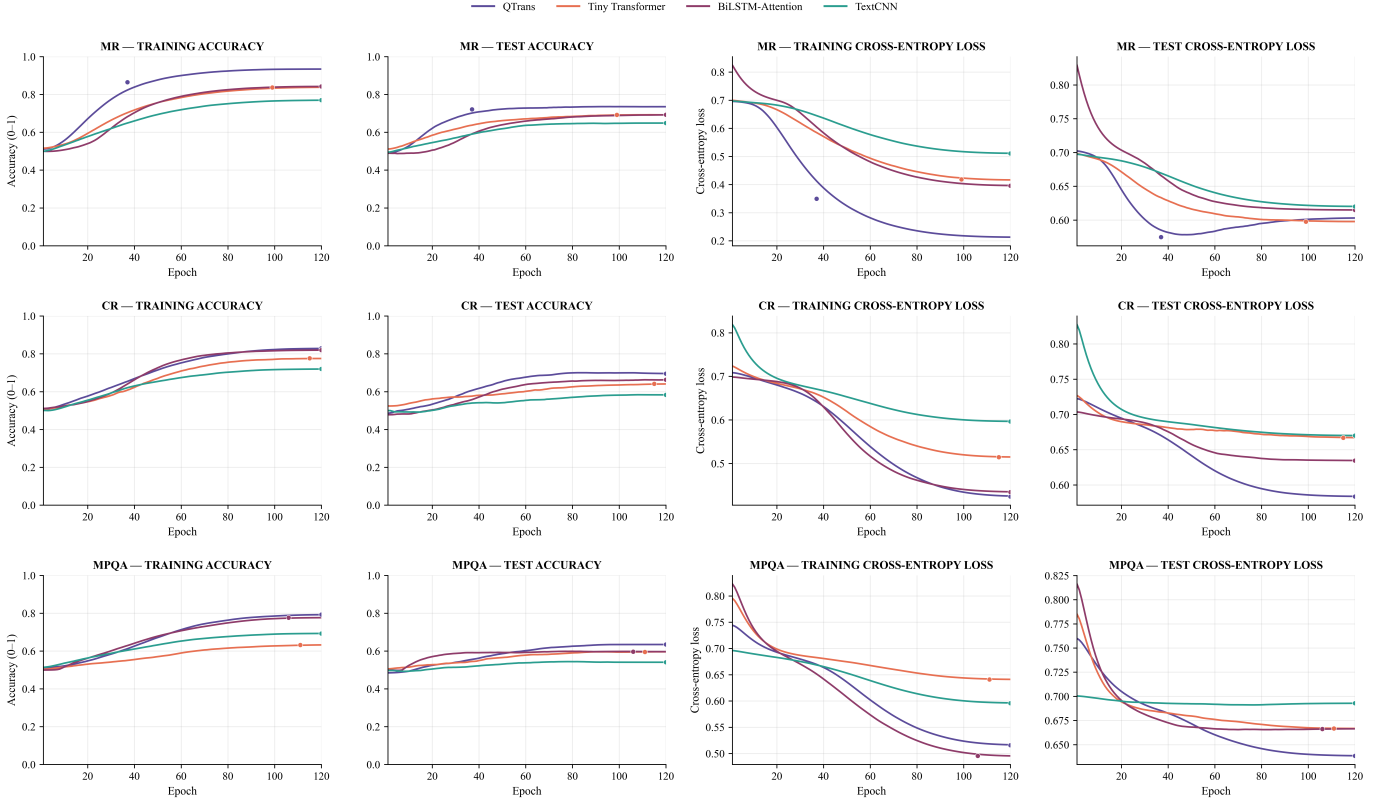}
\caption{Experiment 2: comparison of QTrans, Tiny Transformer, BiLSTM-Attention, and TextCNN on MR, CR, and MPQA over 120 epochs. The panels show training accuracy, test accuracy, training cross-entropy loss, and test cross-entropy loss. Curves use EMA-15 only for visualization, and the dots denote validation-selected checkpoints.}
\label{fig:model_comparison_experiment}
\end{figure*}

\begin{table}[!htbp]
\centering
\caption{Validation-selected test accuracy and macro-averaged F1 score (\%).}
\label{tab:model_comparison_results}
\begin{tabular}{llrr}
\toprule
Dataset & Model & Test accuracy & Macro-F1 \\
\midrule
MR & QTrans & \textbf{72.13} & \textbf{70.12} \\
   & Tiny Transformer & 69.19 & 69.18 \\
   & BiLSTM-Attention & 69.25 & 69.23 \\
   & TextCNN & 64.88 & 64.87 \\
\midrule
CR & QTrans & \textbf{69.51} & \textbf{69.50} \\
   & Tiny Transformer & 64.15 & 64.05 \\
   & BiLSTM-Attention & 66.34 & 66.30 \\
   & TextCNN & 58.29 & 58.29 \\
\midrule
MPQA & QTrans & \textbf{63.45} & \textbf{63.31} \\
     & Tiny Transformer & 59.52 & 59.21 \\
     & BiLSTM-Attention & 59.66 & 59.60 \\
     & TextCNN & 54.06 & 53.76 \\
\bottomrule
\end{tabular}
\end{table}

The comparative results can be analyzed from the following three perspectives.

\begin{itemize}
    \item In terms of test accuracy, QTrans obtains $72.13\%$, $69.51\%$, and $63.45\%$ on MR, CR, and MPQA, respectively. Compared with the strongest classical baseline on each dataset, the corresponding improvements calculated from Tab.~\ref{tab:model_comparison_results} are 2.88, 3.17, and 3.79 percentage points.
    \item Regarding macro-averaged F1, QTrans likewise ranks first on all three datasets. This consistency indicates that the accuracy improvements do not arise merely from favoring one sentiment class, a conclusion further supported by the strict $1:1$ balancing protocol.
    \item During training, the training losses of all four models generally decrease, whereas the test losses of several model-dataset combinations rise slightly in the middle or late stages. This is a genuine manifestation of continued training-set fitting without corresponding generalization improvement. Therefore, the reported checkpoint is selected by validation loss, while all 120 epochs are retained rather than deleting or monotonizing the rising intervals.
\end{itemize}

In summary, under the present fixed-seed and parameter-controlled comparison, QTrans achieves higher accuracy and macro-F1 than the three lightweight classical models. Nevertheless, because the formal results are obtained from one predetermined random seed, ``best performance in this experiment'' should not be interpreted as statistically universal superiority.

\subsubsection{Robustness to Embedding Noise}\label{sec:noise_analysis}

To investigate the influence of input perturbations on quantum multi-head self-attention, Gaussian noise is added to the token embedding of each valid position:
\begin{equation}\label{eq:embedding_noise}
\begin{aligned}
\widetilde{\boldsymbol{x}}_i
&=E_T(t_i)+E_P(i)+\boldsymbol{\varepsilon}_i
=\boldsymbol{x}_i+\boldsymbol{\varepsilon}_i,\\
\boldsymbol{\varepsilon}_i
&\sim\mathcal{N}\!(\boldsymbol{0},\sigma^2I_4),
\qquad m_i=1,
\end{aligned}
\end{equation}
where $\sigma\in\{0,0.1,0.3,0.5\}$. At each training step, one of the four noise levels is sampled uniformly. At the end of each epoch, the training and test sets are separately evaluated at every fixed noise level to produce the four curves shown for each metric. Each test value is averaged over five independent noise realizations. The same set of five random seeds is reused across the four noise levels within an epoch to reduce fluctuations unrelated to noise strength. The complete robustness curves are shown in Fig.~\ref{fig:noise_robustness_experiment}, and the EMA-15-smoothed accuracy and cross-entropy loss displayed at epoch 120 are listed in Tab.~\ref{tab:noise_results}.

\begin{figure*}[!htbp]
\centering
\includegraphics[width=\textwidth]{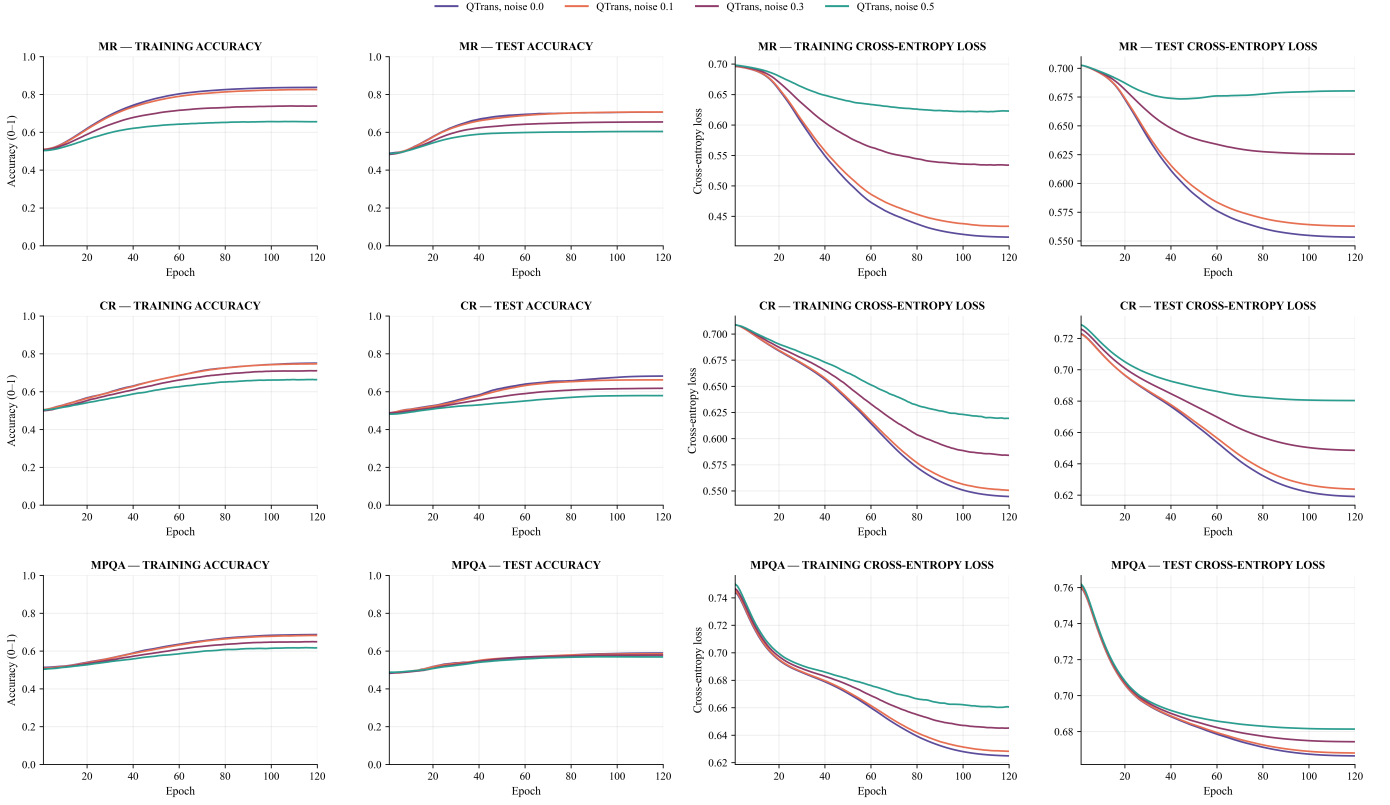}
\caption{Experiment 3: robustness of QTrans under Gaussian embedding noise with $\sigma\in\{0,0.1,0.3,0.5\}$. The panels show training accuracy, test accuracy, training cross-entropy loss, and test cross-entropy loss on MR, CR, and MPQA over 120 epochs. EMA-15 is used only to display the trends.}
\label{fig:noise_robustness_experiment}
\end{figure*}

\begin{table*}[!htbp]
\centering
\caption{EMA-15-smoothed robustness results displayed at epoch 120 under Gaussian embedding noise.}
\label{tab:noise_results}
\small
\setlength{\tabcolsep}{4pt}
\begin{tabular}{lrrrrrrrr}
\toprule
& \multicolumn{2}{c}{$\sigma=0$} & \multicolumn{2}{c}{$\sigma=0.1$} & \multicolumn{2}{c}{$\sigma=0.3$} & \multicolumn{2}{c}{$\sigma=0.5$} \\
\cmidrule(lr){2-3}\cmidrule(lr){4-5}\cmidrule(lr){6-7}\cmidrule(lr){8-9}
Dataset & Acc. (\%) & Loss & Acc. (\%) & Loss & Acc. (\%) & Loss & Acc. (\%) & Loss \\
\midrule
MR   & 70.7 & 0.553 & 70.8 & 0.563 & 65.5 & 0.625 & 60.5 & 0.680 \\
CR   & 68.3 & 0.619 & 66.2 & 0.624 & 61.8 & 0.648 & 57.9 & 0.680 \\
MPQA & 59.2 & 0.666 & 58.6 & 0.668 & 57.9 & 0.674 & 56.9 & 0.681 \\
\bottomrule
\end{tabular}
\end{table*}

The values in Tab.~\ref{tab:noise_results} are read from the EMA-15-smoothed curves at epoch 120 in Fig.~\ref{fig:noise_robustness_experiment}. Accuracies are reported to one decimal place and losses to three decimal places to match the graphical resolution. These values are not the validation-selected checkpoint results reported in Tab.~\ref{tab:model_comparison_results}; consequently, the two tables should not be expected to have identical values at $\sigma=0$.

According to Tab.~\ref{tab:noise_results}, several observations can be derived.

\begin{itemize}
    \item When the noise strength increases from 0 to 0.1, the test accuracies on the three datasets change only moderately. The 0.1-percentage-point increase on MR lies within the graphical and experimental fluctuation and should not be interpreted as a deterministic benefit introduced by noise.
    \item When $\sigma$ reaches 0.3 and 0.5, the accuracies on MR and CR decrease visibly while the losses increase. This trend indicates that medium- and high-strength perturbations have begun to disrupt the token projections and the attention correlations on which the sentiment decision depends.
    \item The variation in MPQA accuracy is comparatively small across noise levels, but its noiseless baseline is already lower than those of MR and CR. Therefore, apparent stability cannot be assessed independently of absolute performance; a smaller decline does not automatically imply stronger effective robustness.
\end{itemize}

Overall, QTrans exhibits a degree of stability under low-strength embedding noise but cannot resist strong input perturbations. It is also noteworthy that this experiment measures the sensitivity of quantum multi-head self-attention to noisy classical representations. It is not equivalent to evaluating gate errors, decoherence, or readout errors on quantum hardware, which should be investigated separately with noisy simulators and real quantum devices.

\section{Conclusion}\label{sec:conclusion}

QTrans is constructed to strengthen contextual relationship modeling in small-scale binary sentiment classification. Independent parameterized quantum circuits generate the query, key, and value measurements, Gaussian-projected coefficients determine token relevance, and a position-wise quantum feed-forward neural network refines the contextualized representation. The quantum multi-head self-attention mechanism and quantum feed-forward neural network are combined with residual connections, layer normalization, and a terminal $[\mathrm{SEP}]$ readout to form an end-to-end trainable quantum-classical architecture. With a four-qubit implementation, QTrans achieves test accuracies of 72.13\%, 69.51\%, and 63.45\% on MR, CR, and MPQA, respectively, exceeding the strongest parameter-matched classical baseline by 2.88, 3.17, and 3.79 percentage points. The macro-averaged F1 results exhibit the same ranking. The ablation experiment further shows that quantum multi-head self-attention improves accuracy consistently across the three datasets, whereas the contribution of the quantum feed-forward neural network depends on the dataset. Under Gaussian embedding noise, QTrans remains comparatively stable at low noise strength but degrades when the perturbation becomes stronger. These results demonstrate the feasibility of integrating quantum multi-head self-attention and a quantum feed-forward neural network into a lightweight Transformer, while the fixed-seed evaluation, small datasets, simulator-based implementation, and absence of quantum hardware noise remain limitations. Further work should therefore examine repeated random seeds, larger and more diverse text corpora, alternative measurement-head configurations, and execution under realistic gate, decoherence, and readout noise.

%
% Each of the commands below will create an unnumbered section with the appropriate heading.
% Remove any sections that are not relevant for your article.
% All sections except suppdata will be removed if the [anonymous] option is used.
% See iopjournal-guidelines.pdf for more information.
%

\endgroup

%\begin{IEEEbiographynophoto}{Xuelong Li} (M'02-SM'07-F'12) is the CTO and Chief Scientist of China Telecom, and he founded the China Telecom Institute of Artificial Intelligence (TeleAI).\end{IEEEbiographynophoto}


\begin{thebibliography}{99}

\bibitem{1} Q. A. Xu et al., “A systematic review of social media-based sentiment analysis: Emerging trends and challenges,” Decision Analytics Journal, vol. 3, p. 100073, 2022.

\bibitem{2} O. Chew et al., “Understanding and mitigating spurious correlations in text classification with neighborhood analysis,” in Findings of the association for computational linguistics: EACL 2024, pp. 1013-1025, 2024.

\bibitem{3} R. Li et al., “Dual graph convolutional networks for aspect-based sentiment analysis,” in Proceedings of the 59th Annual Meeting of the Association for Computational Linguistics and the 11th International Joint Conference on Natural Language Processing (Volume 1: Long Papers), pp. 6319-6329, 2021.

\bibitem{4} L. B. Ilmawan et al., “Negation handling for sentiment analysis task: approaches and performance analysis,” International Journal of Electrical \& Computer Engineering (2088-8708), vol. 14, no. 3, 2024.

\bibitem{5} D. Yin et al., “Sentibert: A transferable transformer-based architecture for compositional sentiment semantics,” in Proceedings of the 58th Annual Meeting of the Association for Computational Linguistics, pp. 3695-3706, 2020.
\bibitem{6} Y. Tian et al., “Enhancing aspect-level sentiment analysis with word dependencies,” in Proceedings of the 16th Conference of the European Chapter of the Association for Computational Linguistics: Main Volume, pp. 3726-3739, 2021.

\bibitem{7} A. Liusie et al., “Analyzing biases to spurious correlations in text classification tasks,” in Proceedings of the 2nd Conference of the Asia-Pacific Chapter of the Association for Computational Linguistics and the 12th International Joint Conference on Natural Language Processing (Volume 2: Short Papers), pp. 78-84, 2022.

\bibitem{8} Y. Aliyu et al., “Sentiment analysis in low-resource settings: A comprehensive review of approaches, languages, and data sources,” IEEE Access, vol. 12, pp. 66883-66909, 2024.

\bibitem{9} P. Xu et al., “Optimizing deeper transformers on small datasets,” arXiv preprint arXiv:2012.15355, 2020.

\bibitem{10} Y. Tian et al., “Aspect-based sentiment analysis with type-aware graph convolutional networks and layer ensemble,” in Proceedings of the 2021 Conference of the North American Chapter of the Association for Computational Linguistics: Human Language Technologies, pp. 2910-2922, 2021.

\bibitem{11} Y. Tay et al., “Long range arena: A benchmark for efficient transformers,” in International Conference on Learning Representations, 2021.

\bibitem{12} T. Lin et al., “A survey of transformers,” AI Open, vol. 3, pp. 111-132, 2022.

\bibitem{13} A. Vaswani et al., “Attention is all you need,” in Proceedings of the 31st International Conference on Neural Information Processing Systems, pp. 6000–6010, 2017.

\bibitem{14} S. Zhang et al., “Alignment attention by matching key and query distributions,” in Proceedings of the 35th International Conference on Neural Information Processing Systems, pp. Article 1030, 2021.

\bibitem{16} P. Michel et al., “Are sixteen heads really better than one?,” in Proceedings of the 33rd International Conference on Neural Information Processing Systems, pp. Article 1257, 2019.

\bibitem{17} T. M. Nguyen et al., “Improving transformer with an admixture of attention heads,” in Proceedings of the 36th International Conference on Neural Information Processing Systems, pp. Article 2026, 2022.

\bibitem{18} H. Peng et al., “A mixture of h-1 heads is better than h heads,” in Proceedings of the 58th Annual Meeting of the Association for Computational Linguistics, pp. 6566-6577, 2020.

\bibitem{19} G. Li et al., “Quantum self-attention neural networks for text classification,” Science China Information Sciences, vol. 67, no. 4, pp. 142501, 2024.

\bibitem{20} M. Schuld et al., “Effect of data encoding on the expressive power of variational quantum-machine-learning models,” Physical Review A, vol. 103, no. 3, pp. 032430, 2021.

\bibitem{21} R. X. Zhao et al., “QKSAN: A quantum kernel self-attention network,” IEEE Transactions on Pattern Analysis and Machine Intelligence, vol. 46, no. 12, pp. 10184-10195, 2024.

\bibitem{22} J. Shi et al., “QSAN: A near-term achievable quantum self-attention network,” IEEE Transactions on Neural Networks and Learning Systems, vol. 36, no. 8, pp. 13995-14008, 2025.

\bibitem{23} R.-X. Zhao et al., “GQHAN: A Grover-inspired quantum hard attention network,” arXiv preprint arXiv:2401.14089, 2024.

\bibitem{24} L. Bischof et al., “Hybrid quantum neural networks show strongly reduced need for free parameters in entity matching,” Scientific Reports, vol. 15, no. 1, pp. 4318, 2025.

\bibitem{25} G. Li et al., “Concentration of data encoding in parameterized quantum circuits,” in Proceedings of the 36th International Conference on Neural Information Processing Systems, pp. Article 1414, 2022.

\bibitem{26} M. Ragone et al., “A Lie algebraic theory of barren plateaus for deep parameterized quantum circuits,” Nature Communications, vol. 15, no. 1, pp. 7172, 2024.

\bibitem{27} R. Xiong et al., “On layer normalization in the transformer architecture,” in Proceedings of the 37th International Conference on Machine Learning, pp. Article 975, 2020.

\bibitem{28} I. Kerenidis et al., “Quantum vision Transformers,” Quantum, vol. 8, pp. 1265, 2024.

\bibitem{29} B. Pang et al., “Seeing stars: Exploiting class relationships for sentiment categorization with respect to rating scales,” in Proceedings of the 43rd Annual Meeting on Association for Computational Linguistics, pp. 115–124, 2005.

\bibitem{30} M. Hu et al., “Mining and summarizing customer reviews,” in Proceedings of the tenth ACM SIGKDD international conference on Knowledge discovery and data mining, pp. 168–177, 2004.



\bibitem{31} T. Wilson et al., “Recognizing contextual polarity in phrase-level sentiment analysis,” in Proceedings of the conference on Human Language Technology and Empirical Methods in Natural Language Processing, pp. 347–354, 2005.

\bibitem{30.1}
S. N. H. Bukhari, Quantum machine learning: Concepts, algorithms, and applications: CRC Press, 2026.

\bibitem{32} J.-J. He et al., “Deepquantum: A pytorch-based software platform for quantum machine learning and photonic quantum computing,” arXiv preprint arXiv:2512.18995, 2025.

\bibitem{33} V. J. B. Jung et al., “Optimizing the deployment of tiny Transformers on low-power MCUs,” IEEE Transactions on Computers, vol. 74, no. 2, pp. 526-541, 2025.

\bibitem{34} J. Deng et al., “Research on sentiment analysis of online public opinion based on RoBERTa–BiLSTM–attention model,” Applied Sciences, vol. 15, no. 4, pp. 2148, 2025.


\bibitem{35} B. Zheng et al., “Sentiment analysis of microblog data based on TextCNN,” in Proceedings of the 2025 2nd International Conference on Computer and Multimedia Technology, pp. 637–643, 2025.

\bibitem{36} I. Loshchilov et al., “Decoupled weight decay regularization,” in International Conference on Learning Representations, 2019.

\bibitem{37} A. M. Smaldone et al., “A hybrid Transformer architecture with a quantized self-attention mechanism applied to molecular generation,” arXiv preprint arXiv:2502.19214, 2025.

\end{thebibliography}
\end{document}